# Adapting Sentence Transformers for the Aviation Domain

Liya Wang[1], Jason Chou[2], David Rouck[3], Alex Tien[4],
Diane M Baumgartner[5]

*The MITRE Corporation, McLean, VA, 22102, United States*

**Learning effective sentence representations is crucial for many Natural Language Processing (NLP) tasks, including semantic search, semantic textual similarity (STS), and clustering. While multiple transformer models have been developed for sentence embedding learning, these models may not perform optimally when dealing with specialized domains like aviation, which has unique characteristics such as technical jargon, abbreviations, and unconventional grammar. Furthermore, the absence of labeled datasets makes it difficult to train models specifically for the aviation domain. To address these challenges, we propose a novel approach for adapting sentence transformers for the aviation domain. Our method is a two-stage process consisting of pre-training followed by fine-tuning. During pre-training, we use Transformers and Sequential Denoising AutoEncoder (TSDAE) with aviation text data as input to improve the initial model performance. Subsequently, we fine-tune our models using a Natural Language Inference (NLI) dataset in the Sentence Bidirectional Encoder Representations from Transformers (SBERT) architecture to mitigate overfitting issues. Experimental results on several downstream tasks show that our adapted sentence transformers significantly outperform general-purpose transformers, demonstrating the effectiveness of our approach in capturing the nuances of the aviation domain. Overall, our work highlights the importance of domain-specific adaptation in developing high-quality NLP solutions for specialized industries like aviation.**

## I. Introduction

In recent years, deep learning has revolutionized the field of Natural Language Processing (NLP) with the development of powerful sentence representation techniques like sentence embeddings. These techniques enable NLP models to capture contextual information about words and their relationships within sentences, making them useful for various artificial intelligence (AI) applications such as semantic search, semantic textual similarity (STS), sentiment analysis, and machine translation.

Two popular approaches for learning sentence embeddings are supervised and unsupervised learning. Supervised learning methods exploit labels for sentence pairs which provide the information about the relation between the sentences, while unsupervised methods rely on large amounts of unannotated data to learn sentence representations without explicit guidance.

Supervised methods include the well-known Sentence Bidirectional Encoder Representations from Transformers (SBERT) [1], which uses Siamese [2] and triplet network structures to derive semantically meaningful sentence embeddings. High-quality sentence embeddings can be derived via supervised training; however, the labeling cost is a major concern in practice, especially for specialized domains. In contrast, unsupervised methods do not need data labels, and have been dominant in sentence embedding learning. There are several types of unsupervised methods, including flow-based, contrastive learning, denoise autoencoder, and prompt-based methods. Flow-based methods include BERT-flow [3] and BERT-whitening [4]. BERT-flow transforms the BERT [5] sentence embedding distribution into a smooth and isotropic Gaussian distribution through normalizing flow [6]. BERT-whitening [4] uses a whitening post-processing method to transform the BERT-based sentence to a standard orthogonal basis while reducing its size.

---

[1]Lead Artificial Intelligence Engineer, Department of Operational Performance
[2]Lead Data Scientist, Department of Operational Performance
[3]Senior Software Engineer, NAS Automation Evolution
[4]Principal Engineer, Department of Operational Performance
[5]Principal Systems Engineer, NAS Future Vision & Research

Contrastive learning methods are popular in sentence embedding learning. The Contrastive Framework for Self-Supervised SEntence Representation Transfer (ConSERT) adopts contrastive learning to fine-tune BERT in an unsupervised way. ConSERT solves the collapse issue [7] of BERT-derived sentence representations to make them more applicable for downstream tasks. Contrastive Tension (CT) [8] treats identical and different sentences as positive and negative pairs and constructs the training objective as a noise-contrastive task between the final layer representations of two independent models, in turn forcing the final layer representations suitable for feature extraction. The Simple Contrastive Learning of Sentence Embeddings (SimCSE) [9] uses contrastive learning to learn sentence embedding from either unlabeled or labeled datasets. SimCSE uses dropout to create identical sentence pairs. Enhanced SimCSE (ESimCSE) [10] further improves the unsupervised learning capability of SimCSE by carefully crafting positive and negative pairs. Difference-based Contrastive Learning for Sentence Embeddings (DiffCSE) [11] learns sentence embeddings from the difference between an original and edited sentence, where the edited sentence is created by stochastically masking out the original sentence and then sampling from a masked language model. Information-aggregated Contrastive learning of Sentence Embeddings (InfoCSE) [12] also derives the sentence embeddings with an additional masked language model task and a well-designed network. Contrastive learning for unsupervised Sentence Embedding with Soft Negative samples (SNCSE) [13], takes the negation of original sentences as soft negative samples and adds Bidirectional Margin Loss (BML) into the traditional contrastive learning framework. The Entity-Aware Contrastive Learning of Sentence Embedding (EASE) [14] learns sentence embeddings via contrastive learning between sentences and their related entities. Contrastive learning method with Prompt-derived Virtual semantic Prototypes (ConPVP) [15] constructs virtual semantic prototypes for each instance, and derives negative prototypes by using the negative form of the prompts. ConPVP uses a prototypical contrastive loss to drive the anchor sentence embedding closer to its corresponding semantic prototypes, and further away from the negative prototypes and the prototypes of other sentences.

Denoise autoencoder and prompt are also be used for unsupervised sentence representation learning. For example, Transformers and Sequential Denoising AutoEncoder (TSDAE) [16], was designed to encode corrupted sentences into fixed-sized embedding vectors and then let the decoder reconstruct the original sentences from this sentence embedding in an unsupervised way. PromptBERT [17] uses prompts to improve BERT sentence embeddings.

It should be mentioned that the models mentioned above were trained on general corpora without considering specific domains, resulting in poor performance when applied directly to domains like aviation. This work seeks to resolve this issue by tailoring pretrained sentence transformers for the aviation domain. Aviation text data are characterized by numerous intricacies like technical jargon, unconventional grammar, and inconsistent abbreviations. In addition, aviation text data have no labels. With those limitations in mind, we designed a two-stage approach comprising pre-training and fine-tuning. We leverage TSDAE during pre-training to enhance the base model's capabilities before refining it further via fine-tuning on the Natural Language Inference (NLI) dataset. By doing so, we ensure better performance than general-purpose pre-trained sentence transformers while minimizing overfitting concerns. Our experiments demonstrate the efficacy of our technique, paving the way for more sophisticated NLP solutions for the aviation sector. We hope that our findings foster further investigation in this promising direction.

The remainder of this paper is organized as follows: Section II gives a short introduction to our input data sources used in the research. Section III provides details of our adaptation modeling process. The results are shown in Section IV. Finally, we conclude in Section V.

## II. Data Sources and Pre-processing

In aviation, various types of text data are accumulated to support safety and daily operations as depicted in Fig. 1. For example, Federal Aviation Administration (FAA) has a Comprehensive Electronic Data Analysis and Reporting (CEDAR) database [18], which provides access to several principal aviation safety data and information sources. The Electronic Occurrence Report (EOR) [19] provides an alert identified by an automated system such as Traffic Analysis and Review Program (TARP) or Operational Error Detection Patch (OEDP) that automatically uploads into the CEDAR tool. The Mandatory Occurrence Report (MOR) [19] reports an occurrence involving air traffic services for which collecting associated safety-related data and conditions is mandatory. Notices to Air Men (NOTAM) [20] are electronic communications to alert aircraft pilots of potential hazards along a flight route or at a location that could affect the safety of the flight. METeorological Aerodrome Report (METAR) [21] reports hourly airport surface weather observations.

These datasets can generally be classified into two main categories based on their linguistic characteristics: domain-specific and everyday language (see Fig. 1). The first group consists of texts written in specialized language often containing technical terms, abbreviations, and acronyms commonly used within the aviation industry, as shown in Table 1. In contrast, the second category encompasses texts that adhere to standard writing conventions without excessive use of jargon or unusual abbreviations.

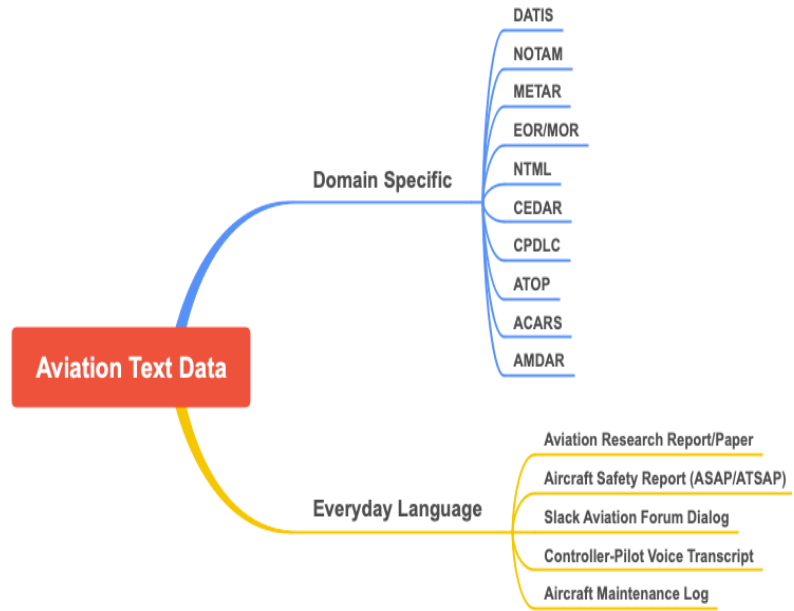

**Fig. 1 Aviation domain text data sources.**

**Table 1 Abbreviated aviation text data example**

```
ANPDAXA
.CHIXCXA 170000
FF KANPXAAT
170000 EDDMZTZW
-ATIS EDDM W SPECI 162359
-EXPECT VECTORS FOR INDEPENDENT PARALLEL ILS APPROACH
-RWY 26R 26L
-NEW ATC SYSTEM IN OPERATION, EXPECT POSSIBLE DELAY
-TRL 60
-RWY 26 LEFT CLSD FM 2100 TILL 0400 UTC ,,
-19001KT
-9999 4000
-RVR RWY26R TDZ P2000 MID 1400 MID P2000 END 1700
-BR BKN024
-T02 DP02
-QNH1020
-
-
-COMMENTS: TG:00
```

Our study focuses on analyzing domain specific texts. Given this focus, we chose the Digital Automatic Terminal Information Service (DATIS) as our primary training data source because it consists exclusively of abbreviated texts from the aviation domain. Since DATIS lacks labels, making supervised fine-tuning impossible, we decided to supplement it with a Natural Language Inference (NLI) dataset. The NLI dataset serves as input during the fine-tuning process, helping us overcome potential overfitting issues. In the subsequent sections, we will describe both datasets in more detail.

## A. Digital Automatic Terminal Information Service (DATIS) Dataset

DATIS systems are widely utilized in busy airports to disseminate information quickly and efficiently [22]. Supported by ARINC [23], DATIS digitally transmits essential Air Traffic Information System (ATIS) notifications, presenting them in an easily comprehensible, written form to flight crews. By doing so, DATIS supports safe and efficient aircraft operation in challenging aeronautical environments.

DATIS communications primarily relay important airport circumstances, such as available landing and departing runways, current meteorological updates, runway closures, taxiway closures, malfunctioning equipment, surface conditions like ice, and other relevant alerts about birds, construction cranes, drones, lasers, etc. This information is combined into a centralized dataset with associated metadata, including timestamps, originating sources, and event dates. This integrated view enables researchers to explore patterns in DATIS usage and assess its effectiveness for various purposes.

Data residing within the MITRE DATIS archive come directly from the Federal Aviation Administration (FAA) via ARINC. Hourly updates take place around the clock, with a one-hour time lag relative to live events. MITRE's database maintains files containing 300 to 400 entries per hour. Table 2 shows examples extracted directly from the logs. This information is a crucial resource for subsequent analysis and investigations related to the use of DATIS information within complex aeronautic contexts.

**Table 2 Raw message data examples**

| Original Message |
| --- |
| ANPDAXA\r\n.ANCATXA 010000\r\nTIS\r\nAD ANC /OS CM2353\r\n- ANC ATIS INFO M 2353Z. 35006KT 2SM -SN BKN017 OVC032 M06/M08 A2929 (TWO NINER TWO NINER) RMK SFC VIS 4. 7R, 7L APPROACHES IN USE.. LANDING RWY 7R, 7L, DEPARTING RWY 7L, 33. NOTAMS... AD WIP SNOW REMOVAL ALL RWYS ALTERNATELY CLSD.. HAZD WX INFO FOR ANC AREA AVBL FM FSS. RWY 7R 5 5 5 2111Z, RWY 7L 5 5 5 2230Z, RWY 33 5 5 5 2112Z. ...ADVS YOU HAVE INFO M.\r\n\n |
| ANPDAXA\r\n.PYCONN2 010000\r\nTIS\r\nAD TNCA/OS CT0000\r\n- \r\nTNCA ARR ATIS T\r\n0000Z\r\nEXP ILS/DME OR VISUAL APPROACH, RWY 11 IN USE.\r\nRUNWAY CONDITION REPORT NOT AVBL.\r\nTRANSITION LEVEL FL40.\r\nNOTAM, ABA VOR/DME FREQ 112.5 MHZ OUT OF SER UFN. \r\nTNCA 010000Z WIND RWY 11 TDZ 110/14KT END 100/14KT VIS 10KM CLD BKN 1800FT T27 DP22 QNH 1012HPA TREND NOSIG=\r\nADZ ON INITIAL CTC YOU HAVE INFO T.\r\n\r\n\n |

To gain an in-depth understanding of the DATIS dataset, we performed exploratory data analysis (EDA) for the year 2022. This allowed us to assess the characteristics of the data, identify patterns and trends, and determine any potential issues that might affect our analysis and interpretation. Through this process, we were able to obtain valuable insights into the properties of the data and develop informed hypotheses about its structure. Our findings from this EDA will serve as a foundation for further analysis and modeling efforts. As shown in Fig. 2, the EDA analysis entailed examining 208 airports featured in the 2022 DATIS dataset. Notably, we observed variations in reporting frequency among the airports, with some updating every 20-30 minutes and others updating their messages irregularly. For example, Hong Kong International Airport did not generate any additional datasets after February 2022. Additionally, there are three primary categories of DATIS messages: combined, arrival, and departure. Smaller airports frequently integrate both arrival and departure details into a single consolidated message, while larger airports, like the Hartsfield–Jackson Atlanta International Airport (ATL), generate separate messages for arrival and departure information.

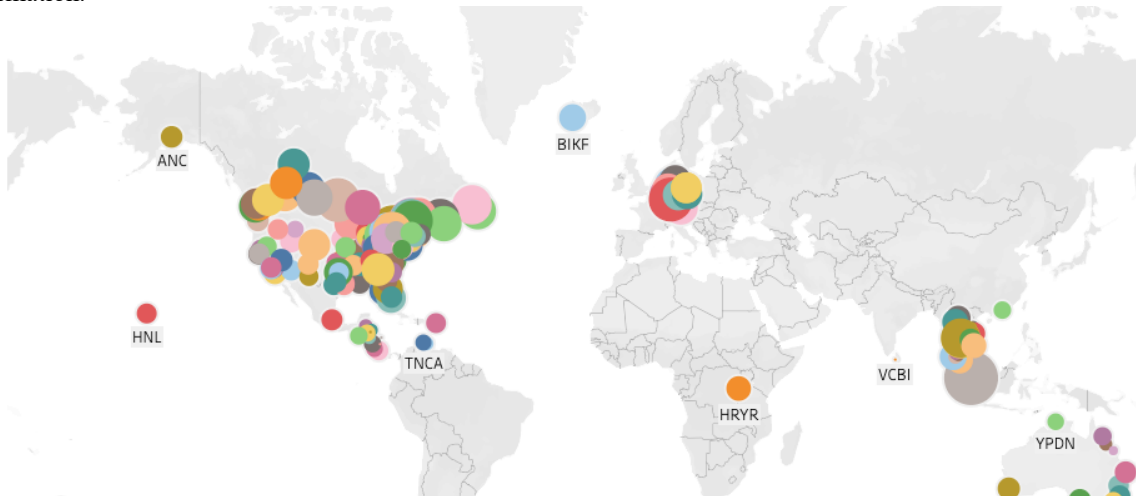

**Fig. 2 DATIS airports in 2022.**

As raw DATIS messages are manually entered by air traffic controllers, they can often contain transcription mistakes. Such errors may result from misspellings, inconsistent abbreviation (e.g., interchangeable use of RY, RWY, or RUNWAY), formatting irregularities (e.g., RWY32L, 18 L, or NOSIG=), improper grammar, extraneous spaces, or omissions. To ensure successful model training using these messages as input, one must thoroughly scrub and cleanse the data prior to analysis.

We developed a set of error correction rules summarized in the green section of Fig. 3. These rules use Python's *re* module [24] to locate specific patterns and make corrections where appropriate. As shown in Table 3, the preprocessing steps lead to cleaner and better organized data, resulting in a significant improvement over the raw messages presented in Table 2. The enhanced quality of the data allows for more accurate and efficient processing and analysis, ultimately leading to better outcomes. These improvements highlight the importance of effective preprocessing techniques when working with text data. After that, we employed the spaCy library [25] to segment DATIS messages into individual sentences, allowing us to gather a corpus consisting of roughly 2,624,012 distinct sentences drawn from the 2022 data files. These sentences constitute our training dataset for future machine learning initiatives.

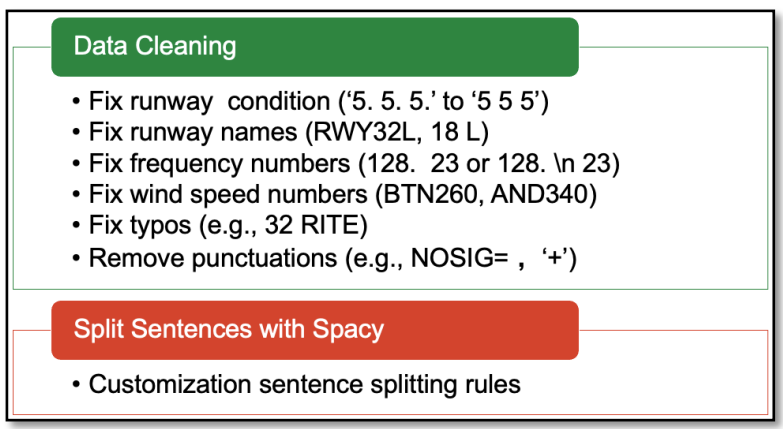

**Fig. 3 DATIS dataset cleaning preprocessing.**

**Table 3 Cleaned message data examples**

| Cleaned Message |
| --- |
| QU ANPDAXA, .ANCATXA 010000, TIS, AD ANC OS CM2353, ANC ATIS INFO M 2353Z. 35006KT 2SM SNOW BKN017 OVC032 M06/M08 A2929 (TWO NINER TWO NINER) RMK SFC VIS 4. 7R, 7L APPROACH IN USE ARRIVING RWY 7R, 7L, DEPARTING RWY 7L, 33. NOTAMS. AD WIP SNOW REMOVAL ALL RWYS ALTERNATELY CLOSED. HAZD WX INFO FOR ANC AREA AVBL FM FSS. RWY 7R 5 5 5 2111Z, RWY 7L 5 5 5 2230Z, RWY 33 5 5 5 2112Z. ADVS YOU HAVE INFO M. |
| QU ANPDAXA, .PYCONN2 010000, TIS, AD TNCA OS CT0000, TNCA ARR ATIS T 0000Z. TNCA 010000Z WIND RWY 11 TDZ 110/14KT END 100/14KT VIS 10KM CLD BKN 1800FT T27 DP22 QNH 1012HPA TREND NOSIG. EXP ILS/DME OR VISUAL APPROACH, RWY 11 IN USE. RWY CONDITION REPORT NOT AVBL. TRANSITION LEVEL FL40. NOTAM, ABA VOR/DME FREQ 112.5 MHZ OUT OF SER UFN. ADZ ON INITIAL CTC YOU HAVE INFO T. |

### B. Natural Language Inference (NLI) Dataset

Natural Language Inference (NLI) involves assessing the truth value of hypotheses based on provided premises. Specifically, NLI categorizes each hypothesis as true (entailment), false (contradiction), or neutral (undetermined). For this study, we obtained the NLI dataset from https://sbert.net/datasets/AllNLI.tsv.gz. This collection contains unions of Stanford Natural Language Inference (SNLI) [26] and MultiNLI [27], resulting in a comprehensive resource with 961,725 records. Having readied the necessary datasets, we proceeded to the next step of model training, detailed in the next section.

### III. Modeling Method

DATIS text data have no labels. With that limitation in mind, we followed the paradigm model training process: pre-training plus fine-tuning (see Fig. 4). During pre-training, we used Transformers and sequential Denoising Auto Encoder (TSDAE) to enhance the base model's capabilities on our aviation dataset. We choose TSDAE because of its relatively better performance reported in [16]. For fine-tuning, we used SBERT to tune sentence transformers with the NLI dataset. This ensures that we achieve better performance than general-purpose pre-trained sentence transformers while minimizing overfitting problems.

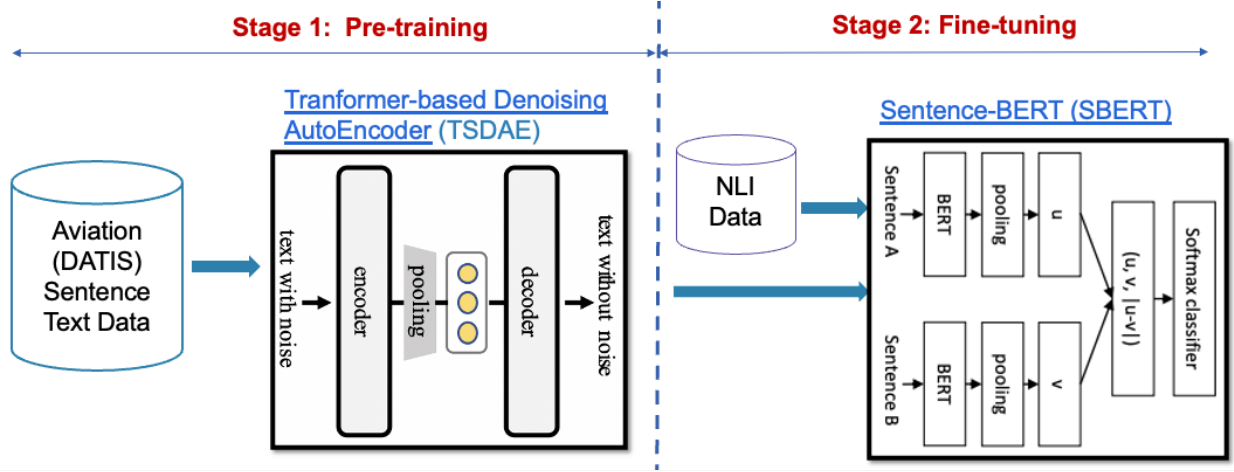

**Fig. 4 Aviation sentence transformer training pipeline.**

**A. TSDAE**

TSDAE is an unsupervised sentence embedding method; it uses a denoise autoencoder [28] as the architecture (see Stage 1 of Fig. 4). During training, TSDAE adds noise to the original sentence, and then feeds it to an encoder which transforms the corrupted sentence into a fixed-sized sentence embedding vector (indicated by yellow in Stage 1 of Fig. 4). Then, the decoder reconstructs the original sentence from this sentence embedding. A good reconstruction denotes that the sentence embedding from the encoder captures the sentence's semantics well. During inference, the encoder is only used for creating sentence embeddings.

TSDAE has modified the conventional encoder-decoder transformer [29]. In TSDAE, the key and value of the cross-attention are both confined to the sentence embedding. Formally, the formulation of the modified cross-attention is:

$$H^{(k)} = Attention(H^{(k-1)}, [S^T], [S^T])$$

$$Attention(Q, K, V) = softmax(\frac{QK^T}{\sqrt{d}})V$$

where $H^{(k)} \in \mathbb{R}^{t \times d}$ represents the decoder hidden state at time step $t$ at the $k$-th layer; $d$ is the dimension size of sentence embedding vector; $[S^T] \in \mathbb{R}^{1 \times d}$ is sentence embedding vector; and $Q, K, V$ are query, key, and value, respectively.

TSDAE determined an effective approach for training based on three components: (1) using deletion with a deletion ratio of 0.6 as the input noise; (2) employing the output from the [CLS] token as a fixed-size sentence representation; and (3) tying encoder and decoder weights during training. This combination has proven to be highly successful in promoting learning.

**B. Sentence-BERT (SBERT)**

The Sentence-BERT (SBERT) [1] model was developed by modifying the pre-trained BERT network [30]. S-BERT involves training the model on a labeled dataset like NLI to generate sentence embeddings that are more accurate and efficient than those produced by standard BERT or RoBERTa [31] models. Specifically, SBERT uses a combination of Siamese and triplet network architecture to create semantically meaningful sentence representations, as shown in Fig. 5. Using SBERT can significantly decrease inference time from approximately 65 hours with BERT or RoBERTa to just 5 seconds without sacrificing accuracy. We fine-tuned the sentence transformers with the labeled NLI dataset to overcome potential overfitting problems resulting from stage 1 of pre-training.

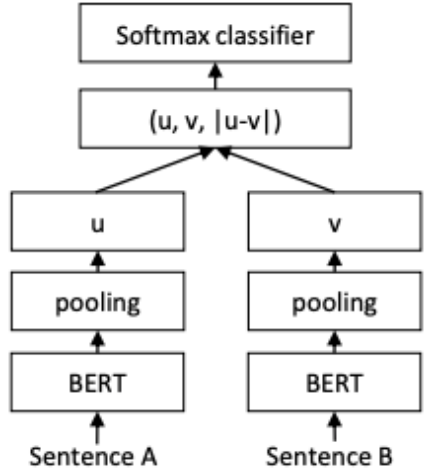

**Fig. 5 SBERT architecture with classification objective function [30].**

## IV. Results

In this section, we present the results of our experiments in applying the aviation sentence transformer to several tasks including STS, clustering, semantic search, and paraphrase mining.

### A. Pretrained Sentence Transformers STS Evaluation

We tested the suitability of pre-trained general-purpose sentence transformer models from the Hugging Face website (https://huggingface.co/sentence-transformers) for use on our selected aviation domain text data. We sought to find the best performing model based on its ability to discern differences between sets of similar or dissimilar sentences. For evaluation purposes, we constructed four test cases in the aviation domain, and computed the cosine similarity score for each sentence pair.

We compiled the resulting scores into Table 4. The bert-base-cased model did not effectively differentiate between sentences in the aviation corpus. As such, it did not meet our requirements, so we excluded it from further consideration. The bert-base-nli-mean-tokens model also fell short of expectations due to its tendency to treat disparate sentences (the Index 2 row in Table 4) with a high cosine similarity score. Conversely, the Index 3 row in Table 4 had highly comparable phrasing, thus providing an ideal test case to measure the capability of the remaining models to generate analogous output. All-MiniLM-L6-v2, all-distilroberta-v1, and all-mpnet-base-v2 also underperformed in this case and were therefore eliminated. Therefore, all-MiniLM-L12-v2 is the final candidate for aviation domain adaptation. The following sections contain additional details about the adaptation experiments and their corresponding results.

**Table 4 Pretrained sentence transformers evaluated on selected aviation text data**

| Index | Sentence1 | Sentence2 | all-MiniLM-L6-v2 | bert-base-nli-mean-tokens | bert-base-cased | all-MiniLM-L12-v2 | all-distilroberta-v1 | all-mpnet-base-v2 |
|---|---|---|---|---|---|---|---|---|
| 0 | NOTAMS. | NOTICE TO AIR MISSIONS. | 0.118 | 0.306 | 0.747 | 0.207 | 0.364 | 0.297 |
| 1 | TDWR OTS. | RWY 2R GS OTS. | 0.543 | 0.788 | 0.882 | 0.580 | 0.660 | 0.572 |
| 2 | HAZDUS WX INFO FOR PHX AREA AVBL ON FSS FREQS. | CLEARANCE FREQUENCY IS 121.9. | 0.201 | 0.574 | 0.914 | 0.311 | 0.299 | 0.355 |
| 3 | BIRD ACTIVITY INVOF ARPT. | WARNING, BIRD ACTIVITY IN VCY OF ARPT | 0.730 | 0.890 | 0.947 | 0.756 | 0.644 | 0.710 |

### B. Experiment Settings

In this section, we describe the training environment used for our model. Table 5 lists our hardware equipment setup. We cloned the entire sentence transformers development package from https://github.com/UKPLab/sentence-transformers. These resources enabled us to effectively train our model and achieve the desired results.

**Table 5 Experimental hardware environment**

| Operation System | Linux |
|---|---|
| CPU | 2xAMD EPYC 7262 8-Core Processor |
| Memory | 250 GB |
| Framework | PyTorch 2.0 |
| GPUs | 4xA100 |

Prior to beginning the training process, we prepared the DATIS training data by formatting each sentence onto a separate line, as needed by the software package being used. We used <sentence-transformers/examples/unsupervised_learning/TSDAE/train_tsdae_from_file.py> as our training script, and we adjusted the training parameters according to those presented in the second column of Table 6. With this configuration, we began the stage 1 training phase.

After completing stage 1, we proceeded to stage 2 of fine-tuning using NLI dataset, using the script <sentence-transformers/examples/training/nli/training_nli_v2.py> and the parameters listed in the third column of Table 6. The script uses the Multiple Negative Ranking Loss strategy [32] where entailment pairs are considered positive while contradictions are treated as hard negatives. Every 10% of the training process, we evaluated the performance of the model on the STS benchmark dataset. When stage 2 was complete, the model was ready to be applied to practical tasks.

**Table 6 Training parameter settings**

| Parameters | Pre-training parameter values | Fine-turning parameter values |
|---|---|---|
| epochs | 1 | 1 |
| weight decay | 1e-5 | 1e-6 |
| scheduler | constant | constant |
| learning rate | 1e-4 | 1e-5 |
| evaluation steps | 500 | 500 |
| save best model | True | True |
| show progress bar | True | True |
| use amp | False | False |
| batch size | 128 | 128 |

### C. Adapted Sentence Transformer STS Evaluation

After completing the two-part training process, we applied the aviation variant of the sentence transformer, named aviation-all-MiniLM-L12-v2, to the same set of text data used in Table 4. The results, listed in Table 7, demonstrate that the adapted aviation-all-MiniLM-L12-v2 model outperforms the general-purpose all-MiniLM-L12-v2. This shows that the adaptation process effectively tailored the model for the domain-specific language patterns prevalent in aviation text.

**Table 7 Adapted model performance comparison**

| Index | Sentence1 | Sentence2 | all-MiniLM-L12-v2 | aviation-all-MiniLM-L12-v2 |
|---|---|---|---|---|
| 0 | NOTAMS. | NOTICE TO AIR MISSIONS. | 0.207 | 0.339 |
| 1 | TDWR OTS. | RWY 2R GS OTS. | 0.580 | 0.655 |
| 2 | HAZDUS WX INFO FOR PHX AREA AVBL ON FSS FREQS. | CLEARANCE FREQUENCY IS 121.9. | 0.311 | 0.288 |
| 3 | BIRD ACTIVITY INVOF ARPT. | WARNING, BIRD ACTIVITY IN VCY OF ARPT | 0.756 | 0.802 |

### D. Clustering Results

We next used aviation-all-MiniLM-L12-v2 model to perform clustering on the DATIS sentences about NOTAM reports from January 1, 2022 to January 9, 2022. The resulting clusters are detailed in Table 8 and visualized using a t-Distributed Stochastic Neighbor Embedding (t-SNE) [26] plot in **Error! Reference source not found.**, which fully demonstrates our adapted sentence transformer was able to identify meaningful patterns in the data. For instance, cluster 0 focuses on runway surface conditions (RSC), while cluster 2 highlights bird activities. Cluster 3 deals with equipment being out of service (OTS), cluster 4 pertains to tower operations, cluster 5 discusses closed taxiways, cluster 6 centers around runway closures, cluster 7 concerns hazardous weather situations, cluster 8 alerts pilots that the tower must call for release from other facilities before allowing them to depart, cluster 9 warns about possible threats from lasers shining into aircraft windows, and cluster 10 provides information on snow. Cluster 1 is a miscellaneous category, containing a broad range of uncommon messages.

**Table 8 Clustering results of sentences**

| Sentence | Cluster id |
|---|---|
| RWY 24 RSC 5 5 5 AT 1335Z, RWY 33 RSC 5 5 5 AT 1335Z | 0 |
| RWY 4L RSC 3 3 3 SLIPPERY WHEN WET AT 1140Z ZULU, RWY 4R RSC 3 3 3 SLIPPERY WHEN WET AT 1203Z ZULU, RWY 3R RSC 3 3 3 SLIPPERY WHEN WET AT 1148Z ZULU | 0 |
| AIRCRAFT EXPECT INTERSECTION DEPARTING FROM RWY THREE ONE LEFT AT NOVEMBER. | 1 |
| WARNING, BIRD ACTIVITY IN VCY OF ARPT, USE EXTREME CAUTION UNAUTHORIZED LASER EVENT AT 2350Z 17 MILES EAST OF BNA AT 1800 FEET GREEN LASER, WEST | 1 |
| WARNING, BIRD ACTIVITY IN VCY OF ARPT, USE EXTREME CAUTION RWY 2L RSC, 3 3 3 AT 1427Z, RWY 2C RSC, 5 5 5 AT 1412Z, RWY 2R RSC, 5 5 5 AT 1355Z, RWY 31 RSC, 5 5 5 AT 1355Z | 2 |
| BIRD HAZARD ADVISORY ON OR IN THE VICINITY OF THE AIRPORT | 2 |
| RWY 10L GS OTS | 3 |
| RWY 35 REILS OTS, WILLOW GROVE ASR SYSTEM OTS, PHL ASDE OTS | 3 |
| TWR FREQ ALL RWYS 123.77. | 4 |
| ALL ACFT IN CONCOURSE AREA ADZ RAMP TWR OF DP AND TRSN | 4 |
| TWY R10 CLOSED, TWY SJ CLOSED | 5 |
| TWY C BETWEEN RWY 28 AND C1 CLOSED | 5 |
| RWY 6/24 CLOSED | 6 |
| RWY 1L/19R CLOSED, RWY 30/12 CLOSED | 6 |
| HAZD WX FOR SLC AREA. | 7 |
| HAZD WX INFO AVBL FM IN FLIGHT ORR FSS | 7 |
| CALL 4 RLS IN EFFECT 4 ACFT ARRIVING SFO, LAS, SLC CONTACT CD 5 PRIOR TO PUSH | 8 |
| CALL 4 RLS IN EFFECT 4 ACFT ARRIVING LAS, SLC, LGB CONTACT CD 5 PRIOR TO PUSH | 8 |
| UNAUTHORIZED LASER ILLUMINATION AT 0015Z 5 SM N OF PIT GREEN LASER | 9 |
| UNATHZD LASER ILMN EVNT 0318Z 5 MILE FINAL, 1900 FEET, BLUE LASER, DIR UNKN | 9 |
| 100 FEET WIDE AND SWEPT FROM CENTER LESS THAN ONE EIGHTH INCH WET SNOW REMAINDER 1 HALF INCH WET SNOW TWYS AND APRONS COVERED WITH 1 HALF INCH WET SNOW 8 FOOT SNOW BANKS OUTER EDGE OF TERMINAL RAMP | 10 |
| ALL RAMPS AND TWYS PATCHY COMPACTED SNOW AND ICE, SANDED | 10 |

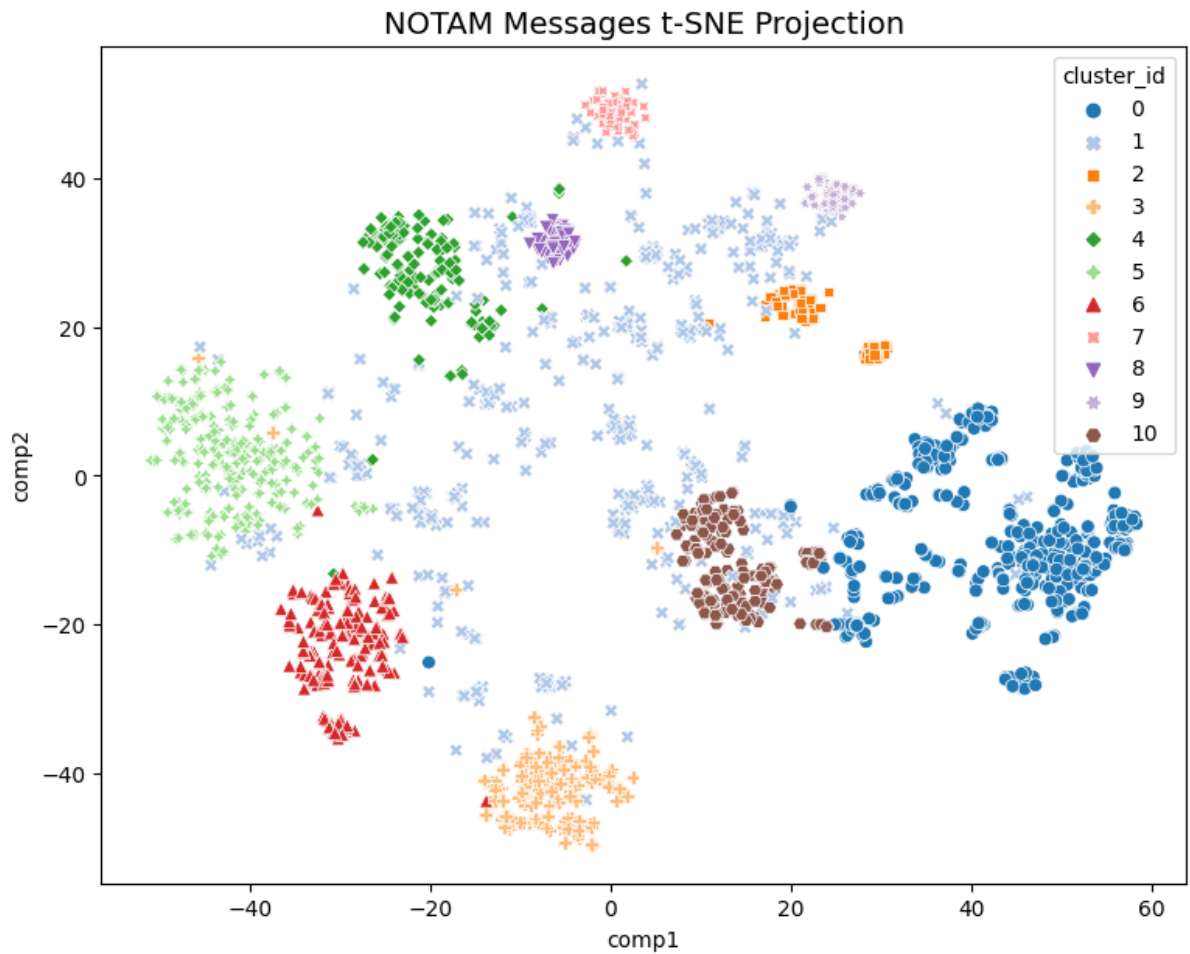

**Fig. 6 t-SNE plot of sentence embedding.**

**E. Semantic Search**

In addition to clustering, we used our newly adapted aviation-all-MiniLM-L12-v2 model to perform semantic searches. By providing a query sentence such as "BIRD ACTIVITY IN THE VICINITY OF THE AIRPORT," the model rapidly identified the ten most similar sentences within the dataset based on their cosine similarity scores; the count column in Table 9 represents how many of the same sentences are included in the searched dataset. Notably, the use of our adapted model allowed for more precise and accurate retrieval of relevant sentences, reflecting its enhanced comprehension of domain-specific language patterns. Furthermore, it underscores the variety of language expressions in the aviation domain.

**Table 9 Semantic search example**

| Query | Sentence | Score | Count |
|---|---|---|---|
| BIRD ACTIVITY IN VCY OF ARPT | BIRD ACTIVITY IN THE VCNTY OF THE ARPT | 0.9744 | 2 |
| BIRD ACTIVITY IN VCY OF ARPT | BIRD ACTIVITY RPTD IN THE VC OF THE ARPT | 0.9661 | 1 |
| BIRD ACTIVITY IN VCY OF ARPT | BIRD ACTIVITY VC OF ARPT | 0.9604 | 2 |
| BIRD ACTIVITY IN VCY OF ARPT | BIRD ACTIVITY VCNTY ARPT | 0.9596 | 2 |
| BIRD ACTIVITY IN VCY OF ARPT | BIRD ACTIVITY VICINITY ARPT | 0.9067 | 2 |
| BIRD ACTIVITY IN VCY OF ARPT | BIRD ACTIVITY VICINITY OF ARPT | 0.8973 | 1 |
| BIRD ACTIVITY IN VCY OF ARPT | BIRD ACTIVITY INVOF ARPT | 0.8806 | 1 |
| BIRD ACTIVITY IN VCY OF ARPT | BIRD ACTIVITY | 0.8518 | 2 |
| BIRD ACTIVITY IN VCY OF ARPT | BIRD ACTIVITY VICINITY DAL ARPT | 0.8353 | 1 |
| BIRD ACTIVITY IN VCY OF ARPT | BIRD ACTIVITY VICINITY ALB ARPT | 0.8196 | 2 |

## F. Paraphrase Mining

To perform paraphrase mining of DATIS messages, we again turned to our aviation-all-MiniLM-L12-v2 model. Unlike previous methods involving brute force comparison, our approach uses the power of the sentence transformer package to quickly and accurately identify duplicate content across larger datasets. Our implementation is guided by the principles introduced in [33]. Table 10 demonstrates the efficacy of this approach, where the scores represent cosine similarity values. When score equals to 1, it means that two messages are identical. This refinement process enables us to streamline the detection of repetitive information while accounting for industry-specific jargon and nuances.

**Table 10 DATIS message paraphrase mining examples**

| Idx1 | Idx2 | Message1 | Message2 | Score |
|---|---|---|---|---|
| 35971 | 35972 | QU ANPDAXA, .CHIXCXA 050345, FF KANPXAAD, 050345 YMENATIS, ATIS YMEN K 050345. WIND: 090/15 MAX XW 15 KTS MAX TW 3 KTS VIS: GT 10KM CLD: FEW030 SCT042 TMP: 27 QNH: 1007. RWY: 17. | QU ANPDAXA, .CHIXCXA 050345, FF KANPXAAD, 050345 YMENATIS, ATIS YMEN K 050345. WIND: 090/15 MAX XW 15 KTS MAX TW 3 KTS VIS: GT 10KM CLD: FEW030 SCT042 TMP: 27 QNH: 1007. RWY: 17. | 1.0000 |
| 54515 | 52842 | QU ANPDAXA, .YQMATXA 070527, TIS, AD CYQM OS CZ0512, CYQM ATIS INFO Z 0500Z. 07006KT 15SM SHSN BKN025 BKN045 M02/M05 A2991. APPROACH RNAV ZULU RWY 29. INFORM MONCTON CENTER ON FREQUENCY 124.4 OF REQUESTED APPROACH ON INITIAL CONTACT. ARRIVING AND DEPARTING RWY 29. RSC RWY 06, RSC 6 6 6 10% ICE, 100% DRY, 100% ICE, VALID AT 2329Z. RSC RWY 29, RSC 6 6 6 100% DRY, 100% DRY, 10% ICE, VALID AT 2335Z. INFORM CYQM ATC ATIS Z. | QU ANPDAXA, .YQMATXA 070106, TIS, AD CYQM OS CV0106, CYQM ATIS INFO V 0100Z. 33007KT 15SM BKN025 BKN040 M00/M04 A2982. APPROACH RNAV ZULU RWY 29. INFORM MONCTON CENTER ON FREQUENCY 124.4 OF REQUESTED APPROACH ON INITIAL CONTACT. ARRIVING AND DEPARTING RWY 29. RSC RWY 06, RSC 6 6 6 10% ICE, 100% DRY, 100% ICE, VALID AT 2329Z. RSC RWY 29, RSC 6 6 6 100% DRY, 100% DRY, 10% ICE, VALID AT 2335Z. INFORM CYQM ATC ATIS V. | 0.9635 |
| 65 | 92 | QU ANPDAXA, .BKKATXA 010010, TIS, AD VTSS OS CA0000, VTSS ARR ATIS A 0012Z. 0000Z WIND 100/4KT VIS 8000M FBL RA CLD FEW CB 1800FT SCT 2000FT BKN 2500FT T23 DP23 QNH 1012HPA TREND NOSIG. RNP 08 12312335 08 5 5 5 100/100/100 NR/NR/NR WET/WET/WET. ADZ CONTROLLER WHEN INITIAL CONTACT YOU HAVE INFO A. | QU ANPDAXA, .LASATXA 010018, TIS, AD LAS OS CY2356, LAS ATIS INFO Y 2356Z. 24009KT 10SM FEW060 13/00 A2951 (TWO NINER FIVE ONE). ILS APPROACH RWY 26L, VISUAL APPROACH IN USE. ARRIVING RWYS 26L AND 19R. DEPARTING RWYS 26R, 19R AND 19L. SIMUL APPROACH TO CROSSING AND PARALLEL RWYS IN USE, CONVERGING RWY OPERATIONS IN EFFECT. NOTAMS. TWY DELTA BETWEEN SIERRA AND MIKE IS RESTRICTED TO MAX WINGSPAN 1 3 5 FEET. HAZD WX INFO AVAILABLE ON HIWAS, FSS FREQ. GC COMBINED ON 121.1, HELICOPTOR CONTROL OPEN ON 118.75. ADVS YOU HAVE INFO Y. | 0.3117 |

## V. Summary

This study describes our novel two-stage training approach utilizing TSDAE and SBERT models to adapt sentence transformers for use on aviation domain text datasets. Experimental evaluation demonstrates significant improvements in various NLP tasks such as STS, clustering, semantic search, and paraphrase mining from methods using general-purpose sentence transformers. Specifically, the adapted model effectively parses DATIS messages, enabling updates regarding weather conditions and other critical landing and departure information to be processed more efficiently. Our experiment results confirm that the adapted model performs well in extracting comprehensible information from text that is dense with abbreviations and domain-specific jargon. Our ongoing research is focused on using the adapted model to support applications that can continuously check for spatial and temporal patterns in reported events to enhance situational awareness and enable proactive mitigation strategies for potential threats to aviation safety. Our proposed adaptation methodology could also be applied to other areas that use a lot of domain-specific language.

## Acknowledgments

The authors thank Dr. Jonathan Hoffman, Dennis Sawyer, Dr. Craig Wanke, Dave Hamrick, Dr. Tom Becher, Mike Robinson, Dr. Lixia Song, Erik Vargo, Matt Yankey, Mahesh Balakrishna, Huang Tang, Shuo Chen, Tao Yu, Michele Ricciardi, and Anahita Imanian of the MITRE Corporation for their support, valuable discussions, and insights.